\pdfoutput=1
\documentclass[11pt]{article}

\usepackage{ACL2023}

\usepackage{times}
\usepackage{latexsym}
\usepackage{amsmath}
\usepackage{amssymb}
\usepackage{graphicx}
\usepackage[frozencache,cachedir=.]{minted}

\usepackage[T1]{fontenc}
\usepackage[utf8]{inputenc}

\usepackage{microtype}

\usepackage{inconsolata}

\title{Ranger: A Toolkit for Effect-Size Based Multi-Task Evaluation}

\author{
  Mete Sertkan\Thanks{All authors contributed equally} \\
  CD-Lab RecSys @ TU Wien \\
  \texttt{\small mete.sertkan@tuwien.ac.at} \\\And
  Sophia Althammer\footnotemark[1] \\
  TU Wien \\
  \texttt{\small sophia.althammer@tuwien.ac.at} \\\And
  Sebastian Hofst{\"a}tter\footnotemark[1] \\
  Cohere\\
  \texttt{\small s.hofstaetter@tuwien.ac.at} \\%\And
  }

\begin{document}
\maketitle
\begin{abstract}
In this paper, we introduce \textit{Ranger} - a toolkit to simplify the utilization of effect-size-based meta-analysis for multi-task evaluation in NLP and IR.
We observed that our communities often face the challenge of aggregating results over incomparable metrics and scenarios, which makes conclusions and take-away messages less reliable.
With \textit{Ranger}, we aim to address this issue by providing a task-agnostic toolkit that combines the effect of a treatment on multiple tasks into one statistical evaluation, allowing for comparison of metrics and computation of an overall summary effect.
Our toolkit produces publication-ready forest plots that enable clear communication of evaluation results over multiple tasks.
Our goal with the ready-to-use \textit{Ranger} toolkit is to promote robust, effect-size based evaluation and improve evaluation standards in the community.
%ill be useful for the community to improve their standards and make evaluation more robust.
We provide two case studies for common IR and NLP settings to highlight \textit{Ranger}'s benefits.
\end{abstract}

% Our main aim is to present an analysis grounded in robust evaluation [51, 60] that does not fall for common problematic shortcuts n IR evaluation like influence of effect sizes [11 , 53 ], relying on too shallow pooled collections [ 2, 33 , 54 ], not accounting for pool bias in old collections [5, 41 , 42 ], and aggregating metrics over different collections which are not comparable [46]. 

\section{Introduction}

We in the NLP (natural language processing) and IR (information retrieval) communities maneuvered ourselves into somewhat of a predicament: We want to evaluate our models on a range of different tasks to make sure they are robust and generalize well. However, this goal is often reached by aggregating results over incomparable metrics and scenarios \cite{thakur2021beir,bowman-dahl-2021-will}. This in turn makes conclusions and take away messages much less reliable than we would like. Other disciplines, such as social and medical sciences have much more robust tools and norms in place to address the challenge of meta-analysis. 

In this paper we present \textit{Ranger} -- a toolkit to facilitate an easy use of effect-size based meta-analysis for multi-task evaluation. \textit{Ranger} produces beautiful, publication-ready forest plots to help everyone in the community to clearly communicate evaluation results over multiple tasks. \textit{Ranger} is written in \texttt{python} and makes use of \texttt{matplotlib}. Thus it will be easy and time-efficient to customize if needed.

\begin{figure*}[t]
    %trim={<left> <lower> <right> <upper>}
   \includegraphics[clip,trim={2.2cm 0.2cm 4.2cm 0.2cm}, width=0.83\textwidth]{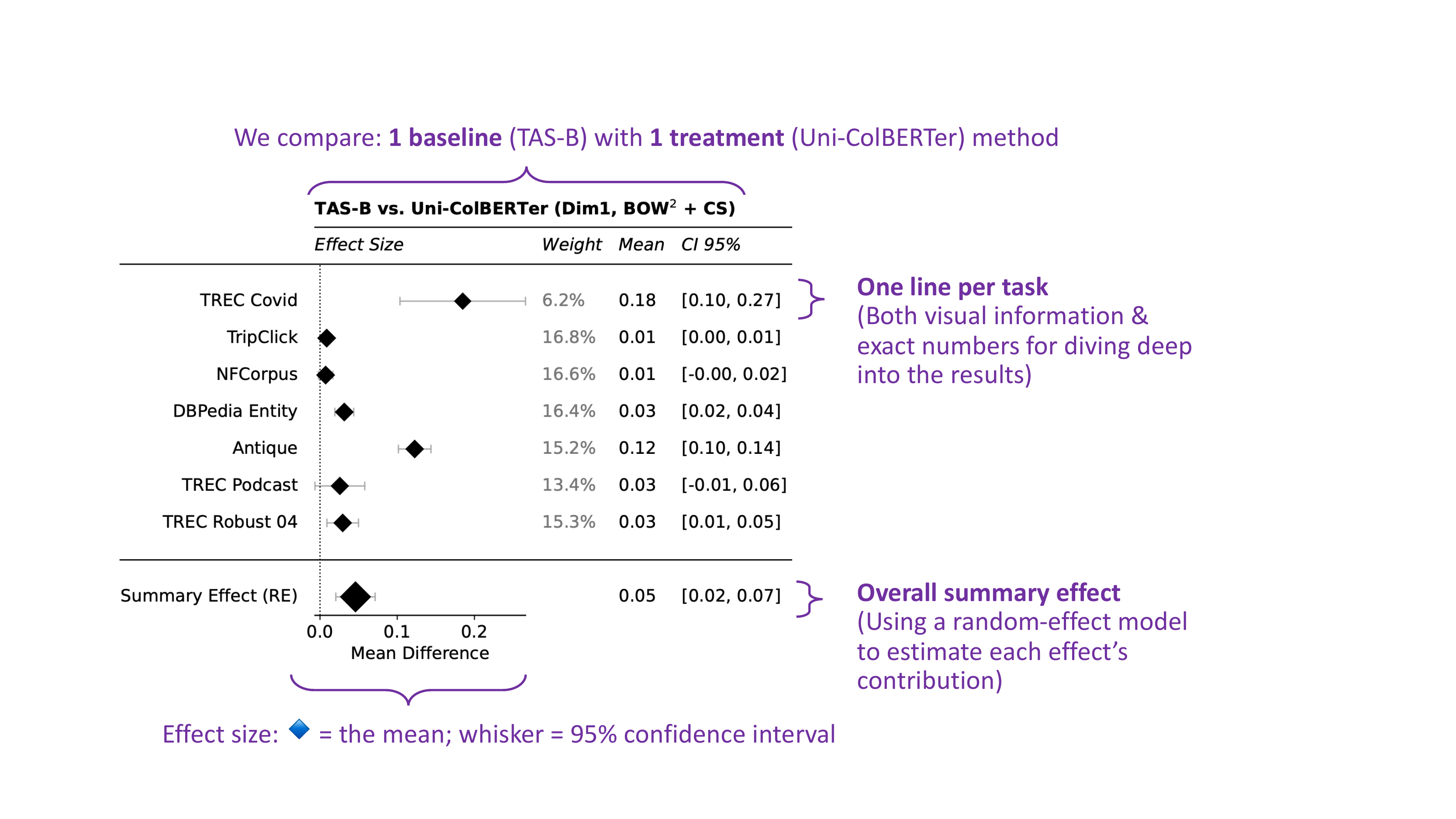}
    \centering
    \vspace{-0.3cm}
    \caption{Example forest plot, with explanations highlighting the output of our \textit{Ranger} toolkit for a multi-task meta-analysis using effect sizes between a control baseline and treatment methods (In this case we use experiments from ColBERTer \cite{Hofstaetter2022_colberter}).}
    \label{fig:hero_figure}
    \vspace{-0.1cm}
\end{figure*}

% combine these advantages more with the usage of Ranger
With the effect-size based meta-analysis \cite{borenstein2009meta} \textit{Ranger} lets you synthesize the effect of a treatment on multiple tasks into one statistical evaluation. Since in meta-analysis the influence of each task on the overall effect is measured with the tasks' effect size, meta-analysis provides a robust evaluation for a suite of tasks with more insights about the influence of one task for the overall benchmark. 
With the effect-size based meta-analysis in \textit{Ranger} one can compare metrics across different tasks which are not comparable over different test sets, like nDCG, where the mean over different test sets holds no meaning.
\textit{Ranger} is not limited to one metric and can be used for all evaluation tasks with metrics, which provide a sample-wise metric for each sample in the test set. \textit{Ranger} can compare effects of treatments across different metrics.
How the effect size is measured, depends on experiment characteristics like the computation of the metrics or the homogeneity of the metrics between the multiple tasks. In order to make \textit{Ranger} applicable to a wide range of multi-task evaluation, \textit{Ranger} offers effect size measurement using the mean differences, the standardized mean difference or the correlation of the metrics. In order to have an aggregated, robust comparison over the whole benchmark, \textit{Ranger} computes an overall combined summary effect for the multi-task evaluation. Since these statistical analysis are rather hard to interpret by only looking at the numbers, \textit{Ranger} includes clear visualization of the meta-analysis comprised in a forest plot as in Figure \ref{fig:hero_figure}.

In order to promote robust, effect-size based evaluation of multi-task benchmarks we open source the ready-to-use toolkit at: \\ \url{https://github.com/MeteSertkan/ranger}

\section{Related Work}

In the last years, increasingly more issues of benchmarking have been discussed in NLP \cite{church-etal-2021-benchmarking,NEURIPS2022_ac4920f4} and IR \cite{craswell2022overview,vorhees2021quality}. \citet{bowman-dahl-2021-will} raise the issue that unreliable and biased systems score disproportionately high on benchmarks in Natural Language Understanding and that constructing adversarial, out-of-distribution test sets also only hides the abilities that the benchmarks should measure.
\citet{bowman-2022-dangers} notice that the now common practices for evaluation lead to unreliable, unrealistically positive claims about the systems.
In IR one common multi-task benchmark is BEIR \citet{thakur2021beir}, which evaluates retrieval models on multiple tasks and domains, however in the evaluation the overall effect of a model is measured by averaging the nDCG scores over each task. As the nDCG score is task dependent and can only be compared within one task, it can not be averaged over different tasks and the mean of the nDCG scores does not hold any meaning. 
Thus there is the urgent need in the NLP and IR community for a robust, synthesized statistical evaluation over multiple tasks, which is able to aggregate scores, which are not comparable, to an overall effect. 

To address these needs \citet{soboroff2018meta} propose effect-size based meta-analysis with the use case of evaluation in IR for a robust, aggregated evaluation over multiple tasks.
Similarily, \citet{NEURIPS2022_ac4920f4} propose a new method for ranking systems based on their performance across different tasks based on theories of social choice.

%
% aggregation for one task: not average but pairing of single test queries 
%\cite{peyrard-etal-2021-better}
%
% meta paper Evaluation Examples Are Not Equally Informative: How Should That Change NLP Leaderboards?
%\cite{rodriguez-etal-2021-change}
In NLP and IR research exist numerous single evaluation tools \cite{azzopardi2019cwleval,macavaney2022irmeasures}, however to the best of our knowledge there exists no evaluation tool addressing the needs for robust, synthesized multi-task evaluation based on meta-analysis.

In order to make this robust, effect-size based evaluation easily accessible for a broad range of tasks, we present our \textit{Ranger} toolkit and demonstrate use cases in NLP and IR.

\section{Ranger}

\subsection{Methodology}
Besides analyzing the effects in individual studies, meta-analysis aims to summarize those effects in one statistical synthesis \cite{borenstein2009meta}.
Translated to the use case of NLP and IR, meta-analysis is a tool to compare whether a treatment model yields gains over a control model within different data collections and overall \cite{soboroff2018meta}.
A treatment, for example, could be an incremental update to the control model, a new model, or a model trained with additional data; a control model can be considered as the baseline (e.g., current SOTA, etc.) to compare the treatment with.
To conduct a meta-analysis, defining an effect size is necessary.
In this work, we quantify the effect size utilizing the raw mean difference, the standardized mean difference, and the correlation.
In particular, we implement the definitions of those effect sizes as defined by \citet{borenstein2009meta} for paired study designs since, typically, the compared metrics in IR and NLP experiments are obtained by employing treatment and control models on the same collections. 

\paragraph{Raw Mean Difference $D$.}
In IR and NLP experiments, researchers usually obtain performance metrics for every item in a collection.
By comparing the average of these metrics, they can make statements about the relative performance of different models.
Thus, the difference in means is a simple and easy-to-interpret measure of the effect size, as it is on the same scale as the underlying metric.
We compute the raw mean difference $D$ by averaging the pairwise differences between treatment $X_T$ and control metric $X_C$ and use the standard deviation ($S_{diff}$) of the pairwise differences to compute its corresponding variance $V_D$ as follows:
\begin{align}
    \begin{split}
        D & =  \frac{X_T-X_C}{n}, \\
        V_D & = \frac{S^2_{diff}}{n}, \\
    \end{split}
\end{align}
where $n$ is the number of compared pairs.

\paragraph{Standardized Mean Difference $d$.}
Sometimes, we might consider standardizing the mean difference (i.e., transforming it into a ``unitless'' form) to make the effect size comparable and combinable across studies.
For example, if a benchmark computes accuracy differently in its individual collections or employs different ranking metrics.
The standardized mean difference is computed by dividing the raw mean difference $D$ by the within-group standard deviation $S_{within}$ calculated across the treatment and control metrics. 
\begin{equation}
    d =\frac{D}{S_{within}} 
\end{equation}
Having the standard deviation of the pairwise differences $S_{diff}$ and the correlation of the corresponding pairs $r$, we compute $S_{within}$ as follows:
\begin{equation}
    S_{within} =\frac{S_{diff}}{\sqrt{2(1-r)}}
\end{equation}
The variance of standardized mean difference $d$ is 
\begin{equation}
    V_d = (\frac{1}{n} + \frac{d^2}{2n})2(1-r),
\end{equation}
where $n$ is the number of compared pairs.
In small samples, $d$ tends to overestimate the absolute value of the true standardized mean difference $\delta$, which can be corrected by factor $J$ to obtain an unbiased estimate called Hedges' $g$ \cite{hedges1981distribution, borenstein2009meta} and its corresponding variance$V_g$:  
\begin{align}
    \begin{split}
        J & = 1 - \frac{3}{4df - 1}, \\
        g & = J \times d, \\
        V_g & = J^2 \times V_d,
    \end{split}
\end{align}
where $df$ is degrees of freedom
%used to compute $S_{within}$, 
which is $n-1$ in the paired study setting with $n$ number of pairs. 

\paragraph{Correlation $r$.}
Some studies might utilize the correlation coefficient as an evaluation metric, for example, how the output of an introduced model (treatment) correlates with a certain gold standard (control).
In such cases, the correlation coefficient itself can serve as the effect size, and its variance is approximated as follows:
\begin{equation}
    V_r = \frac{(1-r^2)^2}{n-1},
\end{equation}
where $n$ is the sample size. 
Since the variance strongly depends on the correlation, the correlation coefficient is typically converted to Fisher's $z$ scale to conduct a meta-analysis \cite{borenstein2009meta}.
The transformation and corresponding variance is: 
\begin{align}
    \begin{split}
        z & = 0.5 \times ln(\frac{1+r}{1-r}), \\
        V_z & = \frac{1}{n-3}
    \end{split}
\end{align}
As already mentioned, $z$ and $V_z$ are used throughout the meta-analysis; however, for reporting/communication, $z$ metrics are transformed back into the correlation scale using: 
\begin{equation}
    r = \frac{e^{2z}-1}{e^{2z}+1}
\end{equation}
\paragraph{Combined Effect $M^*$.}
After calculating the individual effect sizes ($Y_i$) and corresponding variances ($V_{Y_i}$) for a group of $k$ experiments, the final step in meta-analysis is to merge them into a single summary effect.
As \citet{soboroff2018meta}, we assume heterogeneity, i.e., that the effect size variance varies across the experiments.
Following \cite{soboroff2018meta}, we employ the random-effects model as defined in \cite{borenstein2009meta} to consider the between-study variance $T^2$ for the summary effect computation.
We use the DerSimonian and Laird method \cite{dersimonian2015meta} to estimate $T^2$: 
\begin{align}
    \begin{split}
       T^2 & = \frac{Q-df}{C}, \\
       Q & = \sum^k_{i=1}W_iY^2_i - \frac{(\sum^k_{i=1}W_iY^2_i)^2}{\sum^k_{i=1}W_i},  \\
       df & = k-1, \\
       C & = \sum W_i - \frac{\sum W^2_i}{\sum W_i}. 
   \end{split}
\end{align}
where the weight of the individual experiments $W_i = 1/V_{Y_i}$. We adjust the weights by $T^2$ and compute the weighted average of the individual effect sizes, i.e.,  the summary effect $M^*$, and its corresponding variance $V_{M^*}$ as follows:  
\begin{align}
    \begin{split}
        W^*_i & = \frac{1}{V_{Y_i} + T^2}, \\
        M^* & = \frac{\sum^k_{i=1}W^*_iY_i}{\sum^k_{i=1}W^*_i}, \\
        V_{M^*} & = \frac{1}{\sum^k_{i=1}W^*_i}.
    \end{split}
\end{align}

\paragraph{Confidence Interval (CI).}
We determine the corresponding confidence interval (represented by the lower limit, $LL_Y$, and the upper limit, $UL_Y$) for a given effect size $Y$, which can be the result of an individual experiment ($Y_i$) or the summary effect ($M^*$), as follows:
\begin{align}
    \begin{split}
        SE_Y & = \sqrt{V_Y}, \\
        LL_Y & = Y - Z^{\alpha}\times SE_Y, \\
        UL_Y & = Y + Z^{\alpha} \times SE_Y, 
    \end{split}   
\end{align}
where $SE_Y$ is the standard error, $V_Y$ the variance of the effect size, and $Z^{\alpha}$ the Z-value corresponding to the desired significance level $\alpha$. Given $\alpha$ we compute $Z^{\alpha}$:
\begin{equation}
    Z^{\alpha} = ppf(1-\frac{\alpha}{2}),
\end{equation}
where $ppf()$ is the percent point function (we use scipy.stats.norm.ppf\footnote{\url{https://docs.scipy.org/doc/scipy/reference/generated/scipy.stats.norm.html}}).
For example, $\alpha = 0.05$ yields the 95\% CI of $Y \pm 1.96 \times SE_Y$. 

\paragraph{Forest Plots.}
The meta-analysis results in the individual experiments' effect sizes, a statistical synthesis of them, and their corresponding confidence intervals.
Forest plots are a convenient way of reporting those results, which enables a very intuitive interpretation at one glance.
\textit{Ranger} supports forest plots out of the box, which can easily be customized to one's needs since it is based on \texttt{python} and \texttt{matplotlib}.
We provide an example with explanations in Figure~\ref{fig:hero_figure}.
Effect sizes
%are depicted as diamonds $\blacklozenge$,
and corresponding confidence intervals are depicted as diamonds with whiskers $\vdash\blacklozenge\dashv$.
The size of the diamonds is scaled by the experiments' weights ($W^*_i$).
The dotted vertical line $\vdots$ at zero represents the zero effect.
The observed effect size is not significant when its confidence interval crosses the zero effect line; in other words, we cannot detect the effect size at the given confidence level. 

\subsection{Usage}

%We configured \textit{Ranger} so that it is an easy-to-use off-the-shelf evaluation toolkit, in order to promote the use of effect-size based meta-analysis for multi-task evaluation in NLP and IR. 

We explain the easy usage of \textit{Ranger} along with two examples on classification evaluation of GLUE in NLP and retrieval evaluation of BEIR in IR.

The meta-analysis with \textit{Ranger} requires as input either 1) a text file already containing the sample-wise metrics for each task (in the GLUE example) or 2) a text file containing the retrieval runs and the qrels containing the labels (in the BEIR example).

The paths to the text files for each task are stored in a \texttt{config.yaml} file and read in with the class \texttt{ClassificationLocationConfig} or \texttt{RetrievalLocationConfig}. The entry point for loading the data and possibly computing metrics is \texttt{load\textunderscore and\textunderscore compute\textunderscore metrics(name, measure, config)}.
%\begin{minted}{python}
%conf = ClassificationLocationConfig("./config.yaml")
%c = load_and_compute_metrics("bert", conf)
%t = load_and_compute_metrics("distilbert", conf)
%\end{minted}
Having the treatment and control data, we can analyze the effects and compute effect sizes:
%\vspace{0.4cm}
%\hline
%{\small
\begin{minted}{python}
from ranger.metric_containers import
AggregatedPairedMetrics, AggregatedMetrics
from ranger.meta_analysis import
analyze_effects

effects = AggregatedPairedMetrics(
    treatment=t.get_metrics(),
    control=c.get_metrics())
eff_size = analyze_effects(
        list(conf.display_names.values()),
        effects=effects,
        effect_type="SMD")
\end{minted}
%}
%\hline
%\vspace{0.4cm}

Here the \texttt{effect\textunderscore type} variable refers to the type of difference measurement in the meta-analysis as introduced in the previous section. The choice is between Raw Mean differences ("MD"), standardized mean differences ("SMD") or correlation ("CORR").
In order to visualize the effects, \textit{Ranger} produces beautiful forest plots:

\begin{minted}{python}
from ranger.forest_plots import forest_plot

plot = forest_plot(title=title,
experiment_names=list(config.display_names
.values()),
label_x_axis="Standardized Mean Difference",
effect_size=eff_size,
fig_width=8,
fig_height=8)
\end{minted}

\section{Case Study NLP: GLUE benchmark}

\begin{figure}
\centering
  \includegraphics[width=0.5\textwidth ]{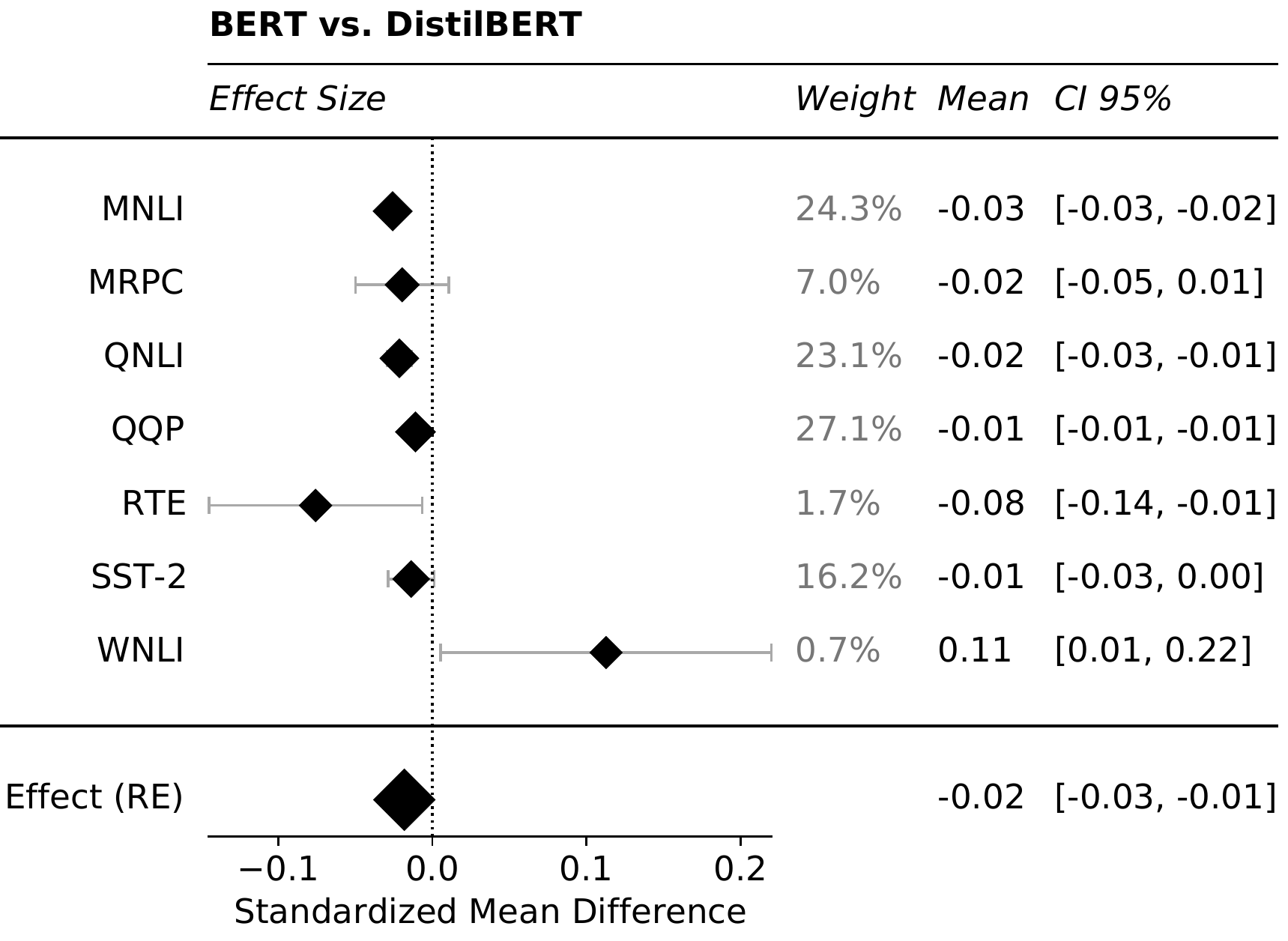}
  \caption{Forest plot of \textit{Ranger} toolkit for tasks of the GLUE benchmark. Comparison in terms of accuracy between BERT and DistilBERT.}
  \label{fig:glue}
  \vspace{-0.5cm}
\end{figure}

In order to demonstrate the usage of the \textit{Ranger} toolkit for various multi-task benchmarks, we conduct an evaluation on the popular General Language Understanding (GLUE) Benchmark \cite{wang-etal-2018-glue}.

We train and compare two classifiers on the GLUE benchmark: one classifier based on BERT \cite{devlin2018bert}, the latter based on a smaller, more efficient transformer model trained on BERT scores, namely DistilBERT \cite{sanh2019distilbert} \footnote{Checkpoints from Huggingface. bert-base-cased for BERT, distilbert-base-cased for DistilBERT.}.

The official evaluation metric for two of the nine tasks (for CoLA and STS-B)  is a correlation-based metric. Since these correlation-based metrics can not be computed sample-wise for each sample in the test set, the effect-size based meta-analysis can not be applied to those metrics and we exclude these two tasks from our evaluation.

We conduct the effect-size based meta-analysis based on the accuracy as metric and use Standardized Mean Difference to measure the effect-size (type in \textit{Ranger} toolkit is 'SMD'). We illustrate the meta-analysis of the BERT and DistilBERT classifier in Figure \ref{fig:glue}. We also publish the Walk-you-Through Jupyter notebook in the \textit{Ranger} toolkit to attain this forest plot for GLUE.

The location of the black diamonds visualizes the effect of the treatment (DistilBERT) compared to the baseline (BERT), whereas the size of the diamonds refers to the weight of this effect in the overall summary effect.
We can see that using DistilBERT as base model for the classifier compared to BERT, has overall effect of a minor decrease in effectiveness.
This behaviour is similar with the results on the MNLI, QNLI, and QQP where we also notice that the confidence intervals are very narrow or even non existent in the forest plot.
For MRPC and SST-2 there is also a negative effect, however the effect is not significant, since the confidence intervals overlap with the baseline performance. 
For RTE and WNLI the effect of using DistilBERT compared to BERT is rather big compared to the summary effect, where for RTE the mean is 8\% lower and for WNLI the mean is 11\% higher than for the BERT classifier. However the large confidence intervals of these tasks indicate the large variability in the effect and thus the weight for taking these effects into account in the summary effect are rather low (0.7\% and 1.7\%).

Overall the summary effect shows that the DistilBERT classifier decreases effectiveness consistently by 2\%. Since the confidence intervals are so narrow for the overall effect and do not overlap with the baseline (BERT classifier), we see that the overall effect is also significant.

\section{Case Study IR: BEIR benchmark}

\begin{figure*}
\centering
  \includegraphics[width=0.8\textwidth]{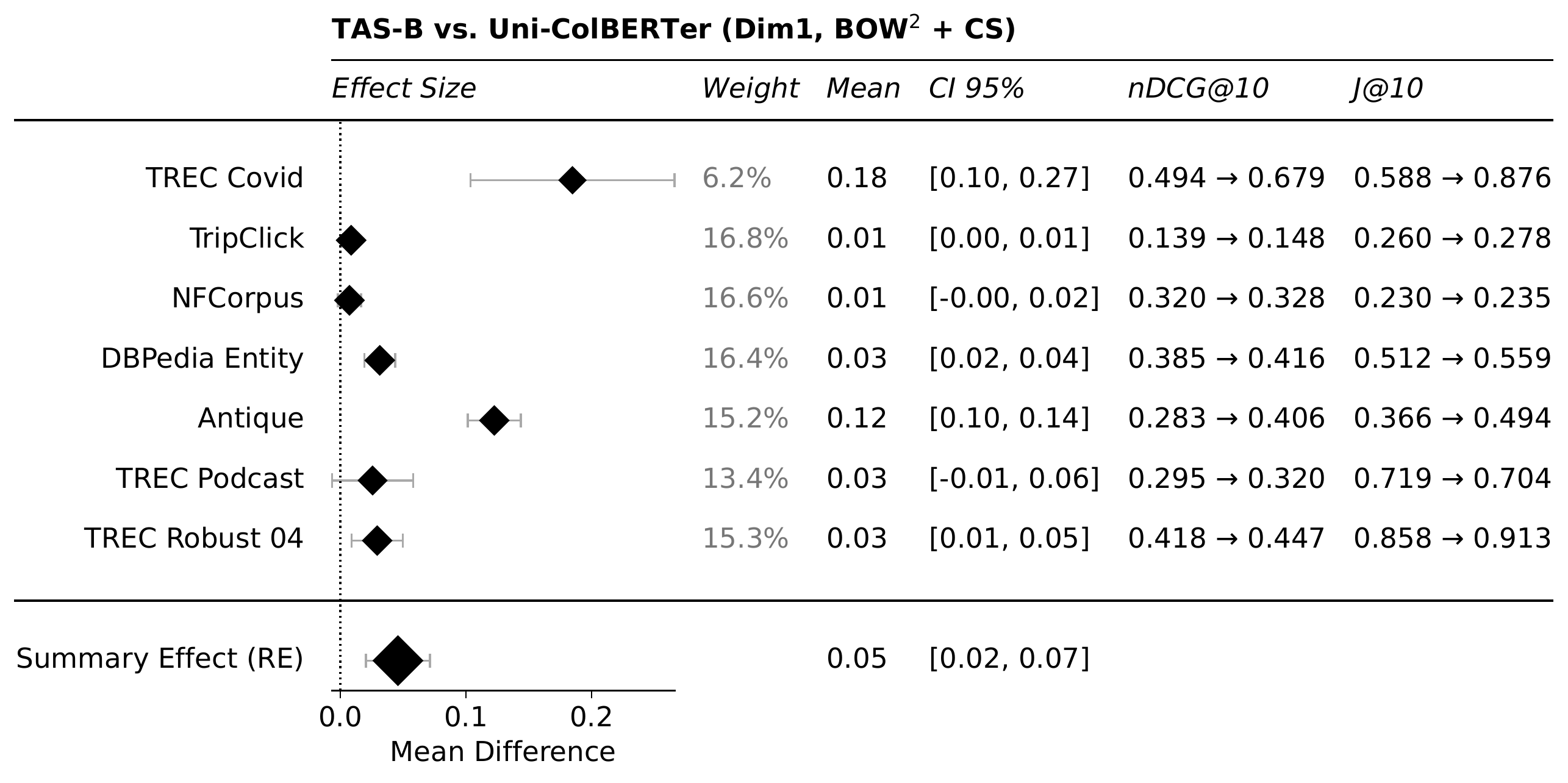}
  \caption{Forest plot of \textit{Ranger} toolkit for tasks of the BEIR benchmark. Comparison in terms of nDCG@10 between TAS-B and ColBERTer.}
  \label{fig:beir}
  \vspace{-0.5cm}
\end{figure*}

Especially in IR evaluation, where it is common to evaluate multiple tasks with metrics, which are not comparable over different tasks \cite{thakur2021beir}, we see a great benefit of using \textit{Ranger} to aggregate the results of multiple tasks into one comparable statistical analysis.
Thus we demonstrate the case study of using the \textit{Ranger} toolkit for evaluation on commonly used IR collections, including a subset of the BEIR benchmark \cite{thakur2021beir}. We presented this study originally as part of \cite{Hofstaetter2022_colberter}, and the \textit{Ranger} toolkit is a direct descendent of these initial experiments.

We select tasks, which either 1) were annotated at a TREC \footnote{https://trec.nist.gov/} track and thus contain high quality judgements, 2) were annotated according to the Cranfield paradigm \cite{cleverdon1997cranfield} or 3) contain a large amount of labels. All collections are evaluated with the ranking metric nDCG@10.
We compare zero-shot retrieval with TAS-B \cite{Hofstaetter2021_tasb_dense_retrieval} as baseline to retrieval with Uni-ColBERTer \cite{Hofstaetter2022_colberter} as treatment.

We conduct a meta-analysis of the evaluation results based on nDCG@10 as metric and measure the effect-size with the mean difference (type is 'MD'). The output of the \textit{Ranger} toolkit is illustrated in Figure \ref{fig:beir}. We publish a walk-through Jupyter notebook in the \textit{Ranger} toolkit to attain this forest plot for BEIR benchmark evaluation.

In Figure \ref{fig:beir} the effect size, the weight of the effect on the overall effect as well as the mean and confidence intervals of the effect are visualized. As an extension for IR we also visualize the nDCG@10 performance and J@10 judgement ratio from baseline $\rightarrow$ to treatment.

For NFCorpus and TREC Podcast we see a small positive effect of Uni-ColBERTer compared to TAS-B, however the confidence intervals are overlapping with the baseline performance incdicating no clear positive effect on these tasks.
For TripClick, DBPedia Entity and TREC Robust 04 we see a consistent and significant small positive effect with narrow confidence intervals of Uni-ColBERTer and this effect is even greater for Antique and TREC Covid. Notice the great confidence intervals for TREC Covid, since the evaluation of TREC Covid is only based on 50 queries and thus its influence for the overall effect should be and is the lowest (6.2\%) among the test sets. 

The judgement ratio J@10 in the left most column shows the percentage of judged documents in the Top 10 of retrieved results. Analyzing the judgement ratio one can also get an understanding of how reliable the evaluation results are and how comparable the results of the two different retrieval models are, since a high difference in judgement ratio could indicate lower comparability of the two models with the respective test set.

Overall the summary effect of Uni-ColBERTer compared to TAS-B is consistent and significantly positive, increasing effectiveness by 0.05.

\section{Conclusion}
We presented \textit{Ranger} -- a task-agnostic toolkit for easy-to-use meta-analysis to evaluate multiple tasks. We described the theoretical basis on which we built our toolkit; the implementation and usage; and furthermore we provide two cases studies for common IR and NLP settings to highlight capabilities of \textit{Ranger}.
We do not claim to have all the answers, nor that using \textit{Ranger} will solve all your multi-task evaluation problems. Nevertheless, we hope that \textit{Ranger} is useful for the community to improve multi-task experimentation and make its evaluation more robust.

{\footnotesize\paragraph{\textbf{Acknowledgements}} This work is supported by the Christian Doppler Research Association (CDG) and has received funding from the EU Horizon 2020 ITN/ETN project on Domain Specific Systems for Information Extraction and Retrieval (H2020-EU.1.3.1., ID: 860721).}

% Entries for the entire Anthology, followed by custom entries
\bibliography{anthology,custom}
\bibliographystyle{acl_natbib}

\appendix

\end{document}